\documentclass[11pt,a4paper]{article}
\usepackage[hyperref]{acl2020}

\usepackage[utf8]{inputenc} 
\usepackage[T1]{fontenc}    
\usepackage{hyperref}       
\usepackage{url}            
\usepackage{booktabs}       
\usepackage{amsfonts}       
\usepackage{nicefrac}       
\usepackage{microtype}      
\aclfinalcopy

\usepackage{hyperref}
\usepackage{url}
\usepackage{times}
\usepackage{latexsym}
\usepackage{amsmath}
\usepackage{amsfonts}
\usepackage{booktabs}
\usepackage{bbm}
\usepackage{graphicx}
\graphicspath{ {Figure/} }
\usepackage{url}
\usepackage{xcolor}
\usepackage{multirow}

\newcommand{\eg}{{e.g.}}

\DeclareMathOperator*{\argmax}{\arg\!\max}

\title{Recurrent Chunking Mechanisms for \\ Long-Text  Machine Reading Comprehension}

\author{
   Hongyu Gong$^1$\thanks{\quad The work was performed during an internship at Tencent AI Lab, Bellevue, WA, USA.} \quad Yelong Shen$^2$\thanks{\quad The work was performed when Yelong Shen was at Tencent AI Lab, Bellevue, WA, USA.} \quad Dian Yu$^3$ \quad Jianshu Chen$^3$ \quad Dong Yu$^3$ \\
   $^1$University of Illinois at Urbana-Champaign, IL, USA\\ \quad $^2$Microsoft Dynamics 365 AI, Redmond, WA, USA \\ $^3$Tencent AI Lab, Bellevue, WA, USA \\
   \texttt{hgong6@illinois.edu}\\
   \texttt{yeshe@microsoft.com} \\
   \texttt{\{yudian,jianshuchen,dyu\}@tencent.com}
   }
   
\begin{document}

\maketitle

\begin{abstract}
In this paper, we study machine reading comprehension (MRC) on long texts, where a model takes as inputs a lengthy document and a question and then extracts a text span from the document as an answer. State-of-the-art models tend to use a pretrained transformer model (e.g., BERT) to encode the joint contextual information of document and question. However, these transformer-based models can only take a fixed-length (e.g., $512$) text as its input. To deal with even longer text inputs, previous approaches usually chunk them into \emph{equally-spaced} segments and predict answers based on each segment independently without considering the information from other segments. As a result, they may form segments that fail to cover the correct answer span or retain insufficient contexts around it, which significantly degrades the performance. Moreover, they are less capable of answering questions that need cross-segment information.

We propose to let a model learn to chunk in a more flexible way via reinforcement learning: a model can decide the next segment that it wants to process in either direction. We also employ recurrent mechanisms to enable information to flow across segments. Experiments on three MRC datasets -- CoQA, QuAC, and TriviaQA -- demonstrate the effectiveness of our proposed recurrent chunking mechanisms: we can obtain segments that are more likely to contain complete answers and at the same time provide sufficient contexts around the ground truth answers for better predictions.
\end{abstract}


\section{Introduction}
\label{sec:intro}
Teaching machines to read, process, and comprehend natural language is a coveted goal of machine reading comprehension (MRC) problems \citep{hermann2015teaching,hill2015goldilocks,rajpurkar2016squad,trischler2017newsqa,zhang2018record,kovcisky2018narrativeqa}. Many existing MRC datasets have a similar task definition: given a document and a question, the goal is to extract a span from the document (in most cases) or instead generate an abstractive answer to answer the question. 


There is a growing trend of building MRC readers~\citep{hu2018read+,xu2019review,yang2019endtoend,keskar2019unifying} based on pre-trained language models~\citep{DBLP:journals/corr/abs-1907-11769,DBLP:journals/corr/abs-1906-08237}, such as GPT~\citep{radfordimproving} and BERT~\citep{bert2018}. These models typically consist of a stack of transformer layers that only allow fixed-length (\eg, $512$) inputs. However, it is often the case that input sequences exceed the length constraint, \eg, documents in the TriviaQA dataset~\cite{triviaQA} contain 2,622 tokens on average. Some conversational MRC datasets such as CoQA~\cite{reddy2018coqa} and QuAC~\cite{choi2018quac} often go beyond the length limit as we may need to incorporate previous questions as well as relatively long documents into the input to answer the current question. 

\begin{figure}[h!]
\centering
\includegraphics[width=0.48\textwidth]{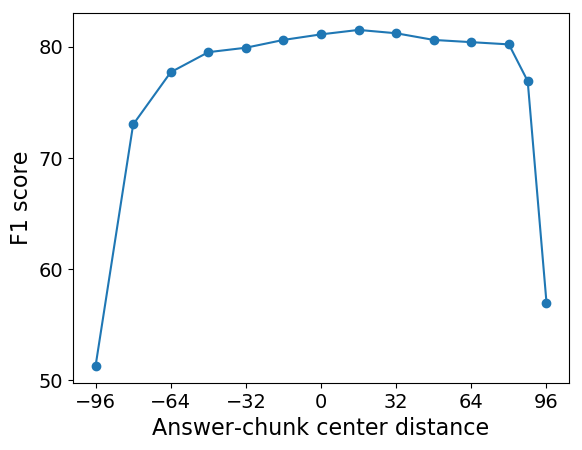}
\caption{The influence of the distance between the center of the answer span and the center of the segment. The test performance (in F1 score) is evaluated on CoQA using a BERT-Large reader. The best performance is achieved when the chunk center coincides with the answer span center. Within the distance of $\pm 80$ (in tokens), while $99\%$ answers are completely covered, the performance degrades as the segment center moves away from the answer center, and the segment contains fewer relevant contexts. When the distance reaches $96$, more than half of the predicted spans are incomplete.}
\label{fig:dist_fscore}
\end{figure}

To deal with long text inputs, a commonly used approach firstly chunks the input text into equally-spaced segments, secondly predicts the answer for each individual segment, and finally ensembles the answers from multiple segments \cite{bert2018}. However, there are two major limitations of this approach: first, a predetermined large stride size for chunking may result in incomplete answers, and we observe that models are more likely to fail when the answer is near the boundaries of a segment, compared to the cases when an answer is in the center of a segment surrounded by richer context (Figure~\ref{fig:dist_fscore}); second, we empirically observe that chunking with a smaller stride size contributes little to (sometimes even hurts) the model performance. A possible explanation is that predicting answer for each segment independently may cause incomparable answer scores across segments. A similar phenomenon is also observed in open-domain question answering tasks~\cite{DBLP:journals/corr/abs-1710-10723}.  

Considering the limitations mentioned above, we propose recurrent chunking mechanisms (RCM) on top of the transformer-based models for MRC tasks. There are two main characteristics of RCM. First, it could let the machine reader learn how to choose the stride size intelligently when reading a lengthy document via reinforcement learning, so it helps prevent extracting incomplete answers from a segment and retain sufficient contexts around the answer. Second, we apply recurrent mechanisms to allow the information to flow across segments. As a result, the model can have access to the global contextual information beyond the current segment.



In the experiments, we evaluate the proposed RCM\footnote{The code is available at \url{https://github.com/HongyuGong/RCM-Question-Answering.git}.} on three MRC datasets: CoQA, QuAC, and TriviaQA. Experimental results demonstrate that RCM leads to consistent performance gains on these benchmarks. Furthermore, it also generates segments that are more likely to cover the entire answer spans and provide richer contextual information around the ground truth answers.

The primary contributions of this work are:
\begin{itemize}
    \item We propose a chunking mechanism for machine reading comprehension to let a model learn to chunk lengthy documents in a more flexible way via reinforcement learning.
    \item We also apply recurrence to allow information transfer between segments so that the model can have knowledge beyond the current segment when selecting answers.
    \item We have performed extensive experiments on three machine reading comprehension datasets: CoQA, QuAC, and TriviaQA. Our approach outperforms two state-of-the-art BERT-based models on different datasets.
\end{itemize}

\section{Method}
\label{sec:method}

\begin{figure*}[h]
\centering
\includegraphics[width=0.97\textwidth]{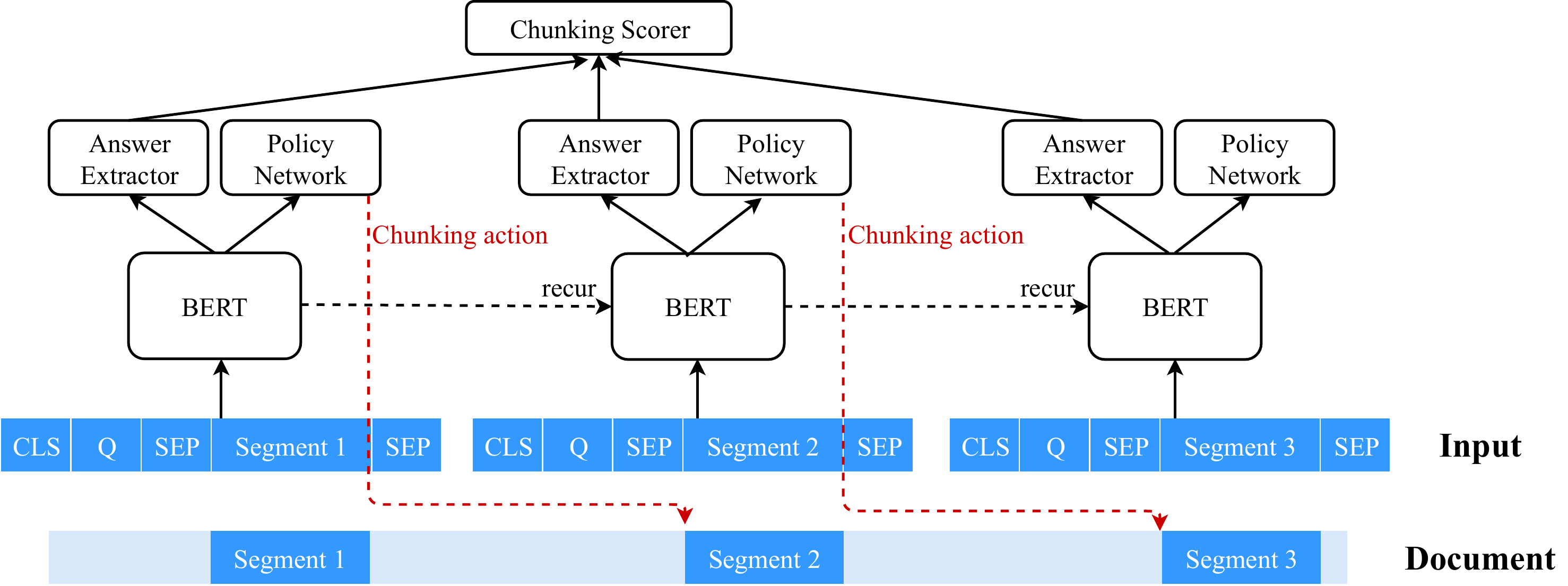}
\caption{BERT generates representations for each input sequence, and recurrence accumulates information over segments. Based on these representations, the answer extractor extracts answers from the current segment, and the policy network takes chunking action and moves to the next segment. Chunking scorer scores each segment by estimating its likelihood of containing an answer and selects answers among predictions from multiple segments.}
\label{fig:system}
\end{figure*}

The proposed recurrent chunking mechanisms (RCM) are built upon the pre-trained BERT models. We will briefly introduce the basic model in Section~\ref{sec:method:model}, and then the RCM approach in Section~\ref{sec:method:recurrent} and~\ref{sec:method:chunk}. More details of our model in training and testing are presented in Sections~\ref{sec:method:training} and ~\ref{sec:method:testing}.


\subsection{Baseline Model}
\label{sec:method:model}
Pre-trained BERT model has been shown to achieve new state-of-the-art performance on many MRC datasets~\citep{bert2018}. Here, we introduce this basic BERT model, which is used as our baseline.
As the maximum input length in BERT is restricted to be $512$, a widely adopted strategy is to chunk a long document into multiple segments with a fixed stride size (i.e., $128$). Following the input format of BERT, the input for each document segment starts with ``CLS'' token, which is followed by question tokens ``Q'' and document segment tokens. We use ``SEP'' token as a separator between the question and the segment. We also append a special ``UNK'' token at the end of the segment to handle unanswerable questions. If a given question is annotated as unanswerable, we mark the ``UNK'' token as the ground truth answer during training. Accordingly in evaluation, if ``UNK'' token is selected by the model from the input segment, we output the answer as “unanswerable”.




\noindent\textbf{Answer Extraction}. Following previous work on extractive machine reading comprehension, we predict the start and the end positions of the answer span in the given document segment. BERT first generates a vector representation $\mathbf{h}_{c,i}$ for each $i$-th token in the $c$-th segment. Given $\mathbf{h}_{c,i}$, the model scores each token in terms of its likelihood of being the start token of the answer span.
\begin{align}
l^{\text{start}}_{c,i} = \mathbf{w}_{s}^{T}\mathbf{h}_{c,i},
\end{align}
\noindent where $\mathbf{w}_{s}$ is the model parameter.
The probability $p^{\text{start}}_{c,i}$ that the answer starts at the $i$-th token is computed by applying the softmax to $l^{\text{start}}_{c,i}$.
\begin{align}
\label{eq:start_prob}
p^{\text{start}}_{c,i} = \mathrm{softmax}(l^{\text{start}}_{c,i})
\end{align}
Likewise, the model scores how likely the answer ends at the $j$-th token in segment $c$ using
\begin{align}
l^{\text{end}}_{c,j} = \mathbf{w}_{e}^{T}\mathbf{h}_{c,j},
\end{align}
where $\bf{w}_{e}$ is the model parameter. The probability of the $j$-th token being the end of the answer (denoted as ${p}^{\text{end}}_{c,j}$) is calculated in a similar manner as Eq. \eqref{eq:start_prob}.

\noindent\textbf{Answer Ensemble}. The baseline model adopts a max-pooling approach to ensemble candidate answers from multiple segments. The answer with the highest probability is selected.


\subsection{Recurrent Mechanisms}
\label{sec:method:recurrent}

The baseline model makes the answer prediction for each document segment independently, which may cause incomparable answer scores across segments due to the lack of document-level information. We propose to use a recurrent layer to propagate the information across different segments and a chunking scorer model to estimate the probability that a segment contains the answer.   

For an input sequence containing the segment $c$, BERT's representation for its first token ``CLS'' is taken as the local representation $v_{c}$ of the segment. The segment representation is further enriched with the representations of previously generated segments via recurrence. We denote the enriched segment representation as $\tilde{{\bf v}}_{c}$:
\begin{align}
\label{eq:general_chunk_rep}
\tilde{{\bf v}}_{c} = f({\bf v}_{c}, \tilde{{\bf v}}_{c-1}),
\end{align}



\noindent where $f(\cdot)$ is the recurrent function. We consider two recurrent mechanisms here: gated recurrence and Long Short Term Memory (LSTM)~\citep{hochreiter1997long} recurrence. 

Gated recurrence is simply a weighted sum of its inputs:
\begin{align}
f_{\text{gated}}({\bf v}_{c}, \tilde{{\bf v}}_{c-1}) = \alpha {\bf v}_{c} + \beta \tilde{{\bf v}}_{c-1},
\end{align}
where $\alpha$ and $\beta$ are coefficients depending on the inputs. We have $\alpha,\beta=\text{softmax}({\bf w}_{r}^{T}[{\bf v}_{c},{\tilde{\bf v}}_{c-1}])$, where ${\bf w}_{r}$ is a model parameter.

The LSTM recurrence, which uses LSTM unit as the recurrence function, takes ${\bf v}_{c}$ as the current input and 
$\tilde{\bf v}_{c-1}$ as the previous hidden state.
\begin{align}
f_{\text{LSTM}}({\bf v}_{c}, \tilde{{\bf v}}_{c-1}) = \text{LSTM}({\bf v}_{c}, \tilde{{\bf v}}_{c-1}).
\end{align}

\noindent\textbf{Chunking Scorer}. Given the enriched segment representation $\tilde{{\bf v}}_{c}$ as input, the chunking scorer produces an scalar $q_{c}$ by:
\begin{align}
\label{eq:chunk_prob}
    q_{c} = \sigma({\bf W}_{c}\tilde{{\bf v}}_{c} + {\bf b}_{c}),
\end{align}
where ${\bf W}_{c}$ and ${\bf b}_{c}$ are model parameters, and $\sigma(\cdot)$ is the sigmoid function. The scalar $q_{c}$ is an estimation of the probability that an answer is included in segment ${c}$. Then, the chunking scorer uses $q_{c}$ to further refine the likelihood of the candidate answers from different segments (see Sections~\ref{sec:method:training} and~\ref{sec:method:testing} for more details on this part of chunking scorer).

\subsection{Learning to Chunk}
\label{sec:method:chunk}

The baseline approach divides a long document into multiple segments with a fixed stride size, from left to right. We will present an approach that could allow the model to choose the stride size flexibly by itself when reading the document. Our motivation, as mentioned in Section~\ref{sec:intro},  is to prevent the answer span from being too close to the segment boundary and covering incomplete answers. 



We formulate the problem of learning-to-chunk under the framework of reinforcement learning. We define the \textbf{state} $s$ of the model to be the segments that a model has processed up to the current segment $c$, i.e., $s=\{1, 2, \ldots, c\}$. The \textbf{action} $a$ is the stride size and direction (forward or backward) the model chooses to move to the next document segment. We define the \textbf{action space} $A$ as a set of strides, e.g., $A = \{-16, 16, 32\}$, where $32$ indicates moving forward with stride size $32$ and $-16$ indicates moving backward with stride size $16$. In this work, we represent the state $s$ with the enriched segment representation $\tilde{{\bf v}}_{c}$.


\noindent\textbf{Chunking Policy}. 
The chunking policy gives the probability ${p}^{\text{act}}(a\, | \,s)$ of taking an action $a$ at the current state $s$, which is modeled by a one-layer feedforward neural network:
\begin{align}
\label{eq:action_prob}
 {p}^{\text{act}}(a\, | \,s) = \text{softmax}({\bf W}_{a}\tilde{\bf{v}}_{c} + {\bf b}_{a}),
\end{align}
where ${\bf W}_{a}$ and ${\bf b}_{a}$ are trainable parameters. 

Fig.~\ref{fig:system} gives an overview of the proposed recurrent chunking mechanisms built upon the BERT model: the chunking policy network takes the enriched segment representation as the input to generate the chunking action, which decides the next segment to be processed. 

\subsection{Training}
\label{sec:method:training}

In the training phase of the recurrent chunking mechanisms, the stride actions of moving to the next segment are sampled according to the probability given by the chunking policy~\citep{sutton2018reinforcement}. Our model generates a sequence of document segments for each question. We train the answer extractor and chunking scorer network with supervised learning, and we train the chunking policy network via reinforcement learning.  

\noindent\textbf{Supervised Learning for Answer Extraction}. Just as the baseline model, we train the answer extraction network via supervised learning. Given a question, the answer extractor classifies whether a word from a document segment is the start or the end of the answer.
The cross-entropy loss can be computed given the ground-truth answer and the predictions of the answer extractor. Suppose that the $i^{*}$-th and $j^{*}$-th tokens are the answer start and end, respectively. The training objective to minimize the following cross-entropy loss, $L_{\text{ans}}$:
\begin{align}
L_{\text{ans}} = -\sum\limits_{c} \log p^{\text{start}}_{c,i*} - \sum\limits_{c}\log p^{\text{end}}_{c,j*},
\end{align}

\begin{table*}[htbp!]
\centering
\resizebox{1.0\textwidth}{!}{
\begin{tabular}{cccc|ccc}
\hline
\multirow{1}{*}{Dataset} 
& \multicolumn{3}{c|}{Train} & \multicolumn{3}{c}{Validation} \\ 
\hline
& Question \# & Avg tokens \# & Max token \# & Question \# & Avg  tokens \# & Max token \# \\
CoQA & 108,647 & 352 & 1,323 & 7,983 & 341 & 1,037 \\
QuAC & 83,568 & 516 & 2,310 & 7,354 & 576 & 2,146 \\
TriviaQA (wiki) & 61,888 & 2,622 & 5,839 & 7,993 & 2,630 & 6,690\\
\hline
\end{tabular}}
\caption{Statistics of the CoQA, QuAC and TriviaQA datasets. We report the number of sub-tokens generated by the BERT tokenizer.}
\label{tab:stat}
\end{table*}

\noindent\textbf{Supervised Learning for Chunking Scorer}. A binary variable $y_{c}$ indicates whether the segment $c$ contains an answer or not. Chunking scorer estimates the probability $q_{c}$ that the segment contains an answer. Similarly, the chunking scorer network can be trained in a supervised manner by minimizing the cross-entropy loss, $L_{\text{cs}}$: 
\begin{align}
    L_{\text{cs}} = -\sum\limits_{c}y_{c}\log q_{c} - \sum\limits_{c} (1- y_{c})\log (1- q_{c}),
\end{align}
where the chunking score $q_{c}$ is given in Eq.~(\ref{eq:chunk_prob}).

\noindent\textbf{Reinforcement Learning for Chunking Policy}. Since the selection of the stride actions is a sequential decision-making process, it is natural to train the chunking policy via reinforcement learning.

First of all, the accumulated reward for taking action $a$ at state $s$ is denoted as $R(s, a)$, which is derived in a recursive manner:
\begin{align}
\label{eq:reward}
    R(s,a) = q_{c}r_{c} + (1-q_{c})R(s',a'),
\end{align}
where $q_{c}$ is the probability that segment $c$ contains an answer as given in Eq.~(\ref{eq:chunk_prob}), and $(s', a')$ denotes the next state-action pair. The value of $r_c$ indicates the probability of the correct answer being extracted from the current segment $c$. The mathematical definition of $r_c$ is given as:   
\begin{align}
    r_{c} = \left\{
    \begin{aligned}
    &p^{\text{start}}_{c,i^{*}} \cdot p^{\text{end}}_{c,j^{*}}, \quad\text{if answer included,}\\
    &0,\qquad\qquad\quad\text{ else.}
    \end{aligned}
    \right.
\end{align}

The first term in Eq.~(\ref{eq:reward}) is the reward of the answer being correctly extracted from the current segment. 
The answer is included in the current segment $c$ with probability $q_{c}$, and thus the first term is weighted by $q_{c}$ in reward $R(s,a)$. The second term in Eq. (\ref{eq:reward}) indicates that $R(s,a)$ also relies on the accumulated reward $R(s',a')$ of the next state when the answer is not available in the current segment.

The chunking policy network can be trained by maximizing the expected accumulated reward (as shown in Eq.~(\ref{eq:reward_j})) through the policy gradient algorithm ~\citep{williams1992simple,sutton2000policy,gong2019reinforcement}. 
\begin{align}
\label{eq:reward_j}
J = \mathbb{E}_{p^{\text{act}}(a\, | \,s)}[R(s,a)].
\end{align}

To be consistent with the notations in answer extraction and chunking scorer modules, we denote the loss function of chunking policy as $L_{\text{cp}}$, which is the negative expected accumulated reward $J$ in Eq.~(\ref{eq:reward_j}): $L_{\text{cp}} = -J$. Thus, the stochastic gradient of $L_{\text{cp}}$ over a mini-batch of data $\mathcal{B}$ is given by:
\begin{align}
\nabla_{}L_{\text{cp}}= -\sum\limits_{(s,a) \in \mathcal{B}}\nabla_{}\log p_{}^{\text{act}}(a\, | \,s)R(s,a),
\end{align}
where $p_{}^{\text{act}}(a\, | \,s)$ is the chunking policy in Eq.~(\ref{eq:action_prob}).

\noindent\textbf{Training procedure}. The overall training loss $L$ is an sum of all three losses: $L=L_{\text{ans}} + L_{\text{cs}} + L_{\text{cp}}.$ In addition, we initialize the bottom representation layers with a pre-trained BERT model and initialize other model parameters randomly. We use the Adam optimizer with peak learning rate $3\times 10^{-5}$ and a linear warming-up schedule. 

\subsection{Testing}
\label{sec:method:testing}
In the testing phase, the model starts from the beginning of the document as its first segment. Later on in state $s$, the model takes the best stride action $a^{*}$ according to the chunking policy:
\begin{align}
a^{*} = \argmax\limits_{a\in A} p^{act}(a\, | \,s)  
\end{align}

After the stride action $a^{*}$ is taken, a new segment is taken from the given document, and so on untill the maximum number of segments $C$ is reached. Now for a document segment $c$, we score its candidate answer spanning from the $i$-th to the $j$-th token by $p_{i,j,c}^{\text{A}}$: 
\begin{align}
\label{eq:ans_joint_prob}
p_{i,j,c}^{\text{A}} = p^{\text{start}}_{c,i} \cdot p^{\text{end}}_{c,j} \cdot q_{c}.
\end{align}

The best answer span $(\bar{i},\bar{j})$ across multiple segments can be obtained by selecting the one with the highest score $p_{i,j,c}^{\text{A}}$. 
\begin{align}
\bar{i},\bar{j} = \argmax\limits_{i\leq j, 1\leq c\leq C} p_{i,j,c}^{\text{A}},
\end{align}
where dynamic programming is used to find ($\bar{i},\bar{j}$) efficiently in linear time.

\section{Experiment}
\label{sec:experiment}

\begin{table*}[htbp!]
\centering
\resizebox{0.96\textwidth}{!}{
\begin{tabular}{lcccc|cccc}
\hline
Dataset & \multicolumn{4}{c|}{CoQA} & \multicolumn{4}{c}{QuAC} \\ 
Max sequence length & 192 & 256 & 384 & 512 & 192 & 256 & 384 & 512 \\ 
\hline

BERT-Large~\cite{bert2018} & 72.8 & 76.2 & 81.0 & 81.4 & 34.5 & 50.6 & 56.7 & 61.5 \\ 
\begin{tabular}[c]{@{}c@{}}Sent-Selector (with previous questions)\end{tabular} & 54.5 & 63.8 & 75.3 & 79.4 & 33.9 & 38.8 & 47.6 & 55.4 \\ 
\begin{tabular}[c]{@{}c@{}}Sent-Selector (only current questions)\end{tabular} & 57.5 & 66.5 & 76.5 & 79.5 & 34.3 & 39.1 & 47.6 & 56.4 \\
\hline
\begin{tabular}[c]{@{}c@{}}BERT-RCM \end{tabular} & & & & & & &  \\
\begin{tabular}[c]{@{}c@{}}- Gated recurrence (no RL chunking) \end{tabular} 
& 74.5 & 78.6 & 81.0 & 81.4 & 48.8 & 51.4 & 56.2 & 61.4 \\
\begin{tabular}[c]{@{}c@{}}- Gated recurrence \end{tabular} 
& \textbf{76.0} & 79.2 & \textbf{81.3} & \textbf{81.8} & 51.6 & 55.2 & 59.9 & \textbf{62.0} \\ 
\begin{tabular}[c]{@{}c@{}}- LSTM recurrence (no RL chunking)\end{tabular} & 74.1 & 78.5 & 81.0 & 81.3 & 49.2 & 51.5 & 56.4 & 61.6 \\
\begin{tabular}[c]{@{}c@{}}- LSTM recurrence\end{tabular} & 75.4 & \textbf{79.5} & \textbf{81.3} & \textbf{81.8} & \textbf{53.9} & \textbf{55.6} & \textbf{60.4} & 61.8 \\
\hline
\end{tabular}}
\caption{Comparison of F1 scores ($\%$) achieved by different algorithms.}
\label{tab:results}
\end{table*}

\subsection{Datasets}
We use three MRC datasets, CoQA~\citep{reddy2018coqa}, QuAC~\citep{choi2018quac} and TriviaQA ~\citep{triviaQA}) in our experiments. 

\noindent(1) \textbf{CoQA}. Answers in the CoQA dataset can be abstractive texts written by annotators.
It is reported that an extractive MRC approach can achieve an upper bound as high as $97.8\%$ in F1 score~\citep{yatskar2018qualitative}. Therefore, We preprocess the CoQA training data and select a text span from the document as the extractive answer that achieves the highest F1 score compared with the given ground truth.\\
\noindent(2) \textbf{QuAC}. All the answers in the QuAC dataset are text spans, which are highlighted by annotators in the given document. \\
\noindent(3) \textbf{TriviaQA}. TriviaQA is a large-scale MRC dataset, containing data from Wikipedia and Web domains. We use its Wikipedia subset in this work. It is reported to be challenging in its variability between questions and documents as well as its requirement of cross-sentence reasoning. Documents in TriviaQA contain more than 2,000 words on average, which is suitable for evaluating the capability of a model to deal with long documents.

The dataset statistics are summarized in Table~\ref{tab:stat}, including the data sizes, the average and maximum number of sub-tokens in documents. 

\subsection{Baselines and Evaluation Metric}

\noindent\textbf{Baselines}. We have two strong baselines based on the pre-trained BERT, which has achieved state-of-the-art performance in a wide range of NLP tasks including machine reading comprehension.\\
\noindent(1) \textsc{BERT-Large model}. It achieves competitive performance on extractive MRC tasks such as SQuAD~\citep{rajpurkar2016squad,rajpurkar2018squad}. It adopts a simple sliding window chunking policy -- moving to the next document segment with a fixed stride size from left to right. We also analyze the performance of the Large BERT model with different stride sizes in training and testing (see Section~\ref{subsec:discuss-bert} for details). The best performance is obtained by setting stride size as $64$ in CoQA and QuAC, and $128$ in TriviaQA. 


\noindent(2) \textsc{Sentence selector}. Given a question, the sentence selector chooses a subset of sentences that are likely to contain an answer. The selected sentences are then concatenated and fed to the BERT-Large model for answer extraction. For conversational datasets CoQA and QuAC, since a question is correlated with its previous questions within the same conversation, we apply the sentence selector to select sentences based on the current question alone or the concatenation of the previous questions and the current question. We only use the current question as the input to the sentence selector for TriviaQA, which does not involve any conversational history. The sentence selector we used in experiments is released by~\newcite{htut2018training}.

\noindent\textbf{Evaluation Metric}. The main evaluation metric is macro-average word-level F1 score. We compare each prediction with the reference answer. Precision is defined by the percentage of predicted answer tokens that appear in the reference answer, and recall is the percentage of reference answer tokens captured in the prediction. F1 score is the harmonic mean of the precision and recall. When multiple reference answers are provided, the maximum F1 score is used for evaluation.

\subsection{Results on CoQA and QuAC}
We first perform experiments on two conversational MRC datasets, CoQA and QuAC.\\
\noindent\textbf{Setting}. 
We perform a set of experiments with different maximum sequence lengths of $192$, $256$, $384$, and $512$. Our model fixes the number of segments read from a document for each question. It generates $4$, $3$, $3$, and $2$ segments under the length limit of $192$, $256$, $384$, and $512$, respectively. 

Considering that questions are highly correlated due to the existence of coreferential mentions across questions, we concatenate each question with as many of its previous questions as possible up to the length limit of $64$ question tokens. The action space of the model strides is set as $[-16, 16, 32, 64, 128]$ for CoQA and  $[-16, 32, 64, 128, 256]$ for QuAC considering that documents in CoQA documents are shorter than those in QuAC. The first segment always starts with the first token of the document, and the model will take stride action after the first segment.

\begin{table*}[htbp!]
\resizebox{1.0\textwidth}{!}{
\begin{tabular}{c|c|c|c|c|c|c|c|c}
\hline
\multicolumn{1}{l|}{Dataset} & \multicolumn{4}{c|}{CoQA} & \multicolumn{4}{c}{QuAC} \\ \hline
\multicolumn{1}{l|}{\# of Doc Tokens} & \textless{}=200 & (200, 300{]} & (300, 400{]} & \textgreater 400 & \textless{}=300 & (300, 450{]} & (450, 600{]} & \textgreater{}600 \\ \hline
\multicolumn{1}{l|}{Percentage (\%)} & 15.3 & 63.3 & 18.9 & 2.5 & 20.5 & 52.0 & 19.7 & 7.8 \\ \hline
\multicolumn{1}{l|}{BERT-Large} & 81.0 & 81.9 & 81.8 & 67.2 & 66.2 & 62.8 & 62.2 & 38.7 \\ 
\hline
\multicolumn{1}{l|}{\multirow{3}{*}{\begin{tabular}[c]{@{}l@{}}BERT-RCM\\ - Gated recurrence\\ - LSTM recurrence\end{tabular}}} &  &  & & & & & & \\
\multicolumn{1}{l|}{} & 81.1 & 82.1 & 82.3 & 74.5 & 66.1 & 62.6 & 63.6 & \textbf{43.2} \\
\multicolumn{1}{l|}{} & 81.1 & 82.0 & 82.3 & \textbf{74.7} & 66.4 & 62.6 & 63.0 & 41.3 \\ \hline
\end{tabular}}
\caption{F1 Score ($\%$) on documents with different numbers of tokens (max sequence length is $512$).}
\label{tab:length_performance}
\end{table*}

\noindent\textbf{Results}. 
In Table~\ref{tab:results}, we present F1 scores achieved by our methods and the baselines. 
The performance of the BERT-Large model drops drastically as the maximum sequence length decreases. We see a drop of $8.6\%$ in F1 score on the CoQA dataset and a drop of $27.0\%$ on the QuAC dataset when the maximum input length decreases from $512$ to $192$. 

Followed by the same BERT-Large reader, the sentence selector baseline that only considers the current question achieves better performance than the selector fed with the concatenation of the current question and its previous questions. The selector with the current question performs well in selecting sentences containing answers from documents. For $90.4\%$ of questions in CoQA and $81.2\%$ of questions in QuAC, the top-ranked $12$ sentences in the documents can include complete answers. However, the selector does not improve upon BERT-Large despite its high precision in sentence selection. This might be because selected sentences do not provide sufficient contexts for a model to identify answers accurately.

Our model with recurrent chunking mechanisms BERT-RCM performs consistently better than both BERT-Large and BERT-Sent-Selector. 
On the CoQA dataset, BERT-RCM with gated recurrence improves upon the BERT-Large model by $3.2\%$, $3\%$, $0.3\%$, and $0.4\%$ with maximum sequence length of $192$, $256$, $284$, and $512$, respectively. The improvement brought by LSTM recurrence and RL chunking is $2.6\%$, $3.3\%$, $0.3\%$, $0.4\%$ on CoQA.  As for the QuAC dataset, gated recurrence combined with RL chunking leads to improvements of $17.1\%$, $4.6\%$, $3.2\%$, $0.5\%$, and LSTM recurrence has gains of $19.4\%$, $5.0\%$, $3.7\%$, $0.3\%$ under different maximum sequence lengths. On the two datasets, the gains of BERT-RCM over BERT-Large are statistically significant at $p=0.05$ with both gated and LSTM recurrence.
We notice that our model is less sensitive to the maximum sequence length, and LSTM recurrence has comparable performance to the gated recurrence.

The gain is more obvious with maximum sequence length $(192, 256, 384)$, and relatively small under the length of $512$. This is perhaps because most document lengths are smaller than $512$ in CoQA and QuAC. Therefore, we report the performance of our proposed method on documents of different lengths in Table~\ref{tab:length_performance}, where the maximum sequence length is set as $512$. We observe that the gain is more obvious on longer documents. For documents with more than 400 words in the CoQA dataset,  RL chunking with gated recurrence has an improvement of $7.3\%$ over BERT-Large, and RL chunking with LSTM recurrence improves F1 score by $7.5\%$. As for QuAC, the improvement of gated recurrence with RL chunking is $4.5\%$, and the improvement of LSTM recurrence is $2.6\%$.

\noindent\textbf{Ablation Analysis}. We further study the effect of recurrence alone without RL chunking here. 
As shown in rows \emph{BERT-Large} and \emph{Gated recurrence (no RL chunking)} in Table \ref{tab:results}, gated recurrence alone can improve F1 score by $2.4\%$, and LSTM recurrence leads to an improvement of $2.3\%$ without RL chunking when the maximum sequence length is $256$. However, we do not observe any improvement when the maximum sequence length is set to $384$ or $512$. 


\begin{table}[htbp!]
\centering
\begin{tabular}{cc}
\hline 
\multicolumn{1}{l|}{Algorithms} & F1 \\
 \hline
\multicolumn{1}{l|}{BERT-Large} & 61.3  \\ 
\multicolumn{1}{l|}{Sent-Selector} & 59.8 \\ 
\hline
\multicolumn{1}{l|}{\multirow{3}{*}{\begin{tabular}[c]{@{}l@{}}BERT-RCM\\ 
- Gated recurrence\\ - LSTM recurrence\end{tabular}}} & \\
\multicolumn{1}{l|}{} & \textbf{62.9}\\ 
\multicolumn{1}{l|}{} & 62.3 \\ 
\hline
\end{tabular}
\caption{F1 score (\%) of different algorithms on the TriviaQA dataset.}
\label{tab:triviaqa}
\end{table} 
\subsection{Results on TriviaQA}
We further evaluate the ability of our model in dealing with extremely long documents on the TriviaQA Wikipedia dataset. 

\begin{table*}[hpbt!]
\centering
\begin{tabular}{l|cccc|cccc}
\toprule
Dataset & \multicolumn{4}{c|}{CoQA} & \multicolumn{4}{c}{QuAC} \\ 
\hline
BERT-Large ~\cite{bert2018} & \multicolumn{4}{c|}{Prediction Stride Size} & \multicolumn{4}{c}{Prediction Stride Size} \\ 
Training Stride Size  & 16 & 32 & 64 & 128 & 16  & 32 & 64 & 128 \\ 
\midrule
16 & 80.8 & 80.9 & 80.8 & 80.7 & 60.6 & 60.7 & 60.7 & 60.8 \\ 
\begin{tabular}[c]{@{}c@{}}  32\end{tabular} & 81.1 & 81.1 & 81.1 & 81.1 & 60.7 & 60.7 & 60.9 & 61.0 \\ 
\begin{tabular}[c]{@{}c@{}} 64\end{tabular} & \textbf {81.4} & \textbf{ 81.4} & \textbf{81.4} & 81.3 & 61.0 & 61.0 & \textbf{61.4} & \textbf{61.4} \\ 
\begin{tabular}[c]{@{}c@{}} 128\end{tabular} & 81.0 & 81.1 & 81.1 & 81.1 & 60.8 & 60.8 & 60.8 & 61.2 \\ 
\bottomrule
\end{tabular}
\caption{F1 score ($\%$) of the BERT-Large model with different training/prediction stride sizes on the CoQA and QuAC datasets.}
\label{tab:stride_size_table}
\end{table*}

\noindent\textbf{Setting}. We set the maximum sequence length as $512$ for all models. The action space of our BERT-RCM model is set to $[-64, 128, 256, 512, 1024]$. The stride sizes are larger than those in CoQA and QuAC, since TriviaQA provides much longer documents. During training, the maximum number of segments our model can extract from a document is set to three in the TriviaQA dataset. Note that our model reads no more than $512 \cdot 3 = 1536$ tokens from these three segments, which are much fewer than the average document length. 

\noindent\textbf{Results}. We filter a small number of questions whose answers cannot be extracted from documents and keep 7,251 questions from a total of 7,993 questions. In Table~\ref{tab:triviaqa}, we present the F1 scores of different algorithms. Compared with BERT-Large, the BERT-RCM model achieves $1.6\%$ gain with gated recurrence and $1\%$ gain with LSTM recurrence. Also, both BERT-RCM and BERT-Large models beat the Sent-Selector model.

\section{Discussion}
\label{sec:discuss}
In this section, we will analyze the performance of the baseline BERT-Large model and our proposed recurrent chunking mechanisms. 
\subsection{Analysis of different Stride Sizes in BERT-Large}
\label{subsec:discuss-bert}
In Table \ref{tab:stride_size_table}, we give an analysis of how the performance varies with different stride sizes in BERT-Large model (the baseline) training and prediction. An interesting observation is that smaller stride size in prediction does not always improve the performance, sometimes even hurts as can be seen on the QuAC dataset. It suggests that BERT-Large performs badly on selecting good answers from multiple chunks. Smaller stride size in model training also leads to worse performance. A possible explanation is that smaller stride size would cause the significant distortion of training data distribution, since the longer question-document pairs produces more training samples than short ones. 

\subsection{Discussions of Recurrent Chunking Mechanisms}
\label{subsec:discuss-rcm}
We now provide an insight into the recurrent mechanisms and chunking policy learned by our proposed model using quantitative analysis. For the clarity of our discussions, we use the following setting on the CoQA and QuAC datasets: the maximum chunk length is set to $256$, and the stride size of BERT-Large model is $128$. 

\textbf{Segment-Hit Rate}. With the ability of chunking policy, BERT-RCM is expected to focus on those document segments that contain an answer. To evaluate how well a model can capture good segments, we use \emph{hit rate}, i.e., the percentage of segments that contain a complete answer among all extracted segments, as evaluation metric. 

\begin{table}[htbp!]
\centering
\begin{tabular}{lc|c}
\toprule
Hit rate  & CoQA & QuAC\\
\midrule
BERT-Large & 54.0 & 34.1 \\
\hline
BERT-RCM & & \\
- Gated recurrence & 73.1 & 44.9 \\
- LSTM recurrence & 79.7 & 42.8 \\
\bottomrule
\end{tabular}
\caption{Comparison of BERT-Large and BERT-RCM on segment-hit rate (\%).}
\label{tab:chunk_hit_rate}
\end{table}

As shown in Table \ref{tab:chunk_hit_rate}, BERT-RCM significantly outperforms BERT-Large, which indicates that the learned chunking policy is more focused on informative segments.

\begin{figure}[h!]
\centering
\includegraphics[width=0.48\textwidth]{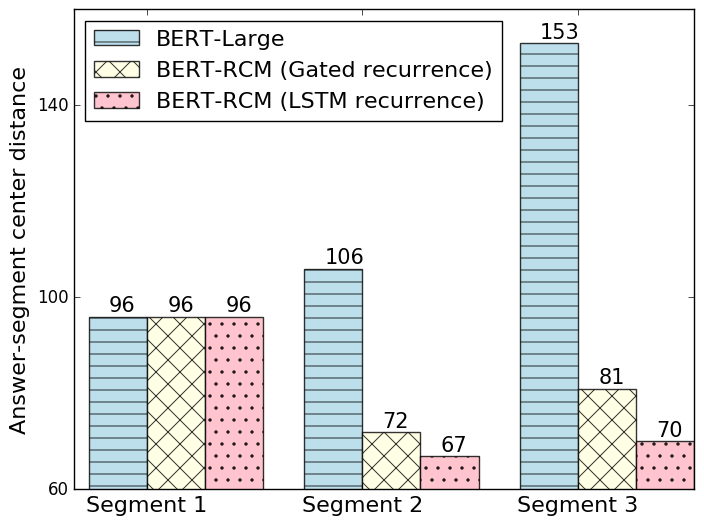}
\caption{The answer-segment center distance.}
\label{fig:chunk_dist_change}
\end{figure}

\begin{figure*}[htbp!]
\centering
\includegraphics[width=1.0\textwidth]{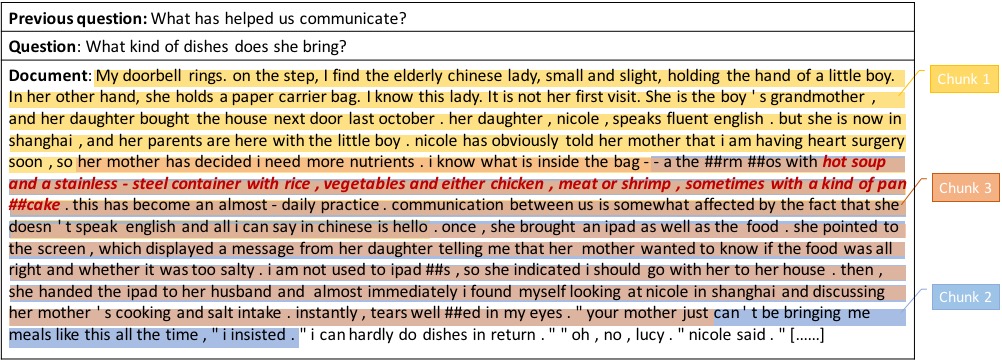}
\caption{Example of generated document segments by BERT-RCM from a CoQA document.}
\label{fig:chunk_example_jump}
\end{figure*}

\textbf{Answer-Chunk Center Distance}. As discussed in Fig.~\ref{fig:dist_fscore}, the answer's position with respect to a document segment is important for answer prediction. When an answer is centered within the document segment, sufficient contexts on both sides help a model make better predictions. In Fig.~\ref{fig:chunk_dist_change}, it presents the averaged center distances of the first three segments generated by BERT-Large and BERT-RCMs on the CoQA validation dataset. 
Since all models start from the beginning of a document in the first segment, their first answer-chunk center distances are the same: $96$ tokens. But for the second and third segments generated by BERT-RCMs, the answer-chunk center distances are much smaller than BERT-Large. 

In this section, we also illustrate the working flow of BERT-RCM with a case study. 

\textbf{Case Study}. We show an example from a CoQA document in Figure \ref{fig:chunk_example_jump} to illustrate the chunking mechanism of our BERT-RCM model with LSTM recurrence. The model starts with the beginning of the document as the first segment, where the answer span is close to its right boundary. The model moves forwards $128$ tokens to include more right contexts and generates the second chunk. The stride size is a bit large since the answer is close to the left boundary of the second segment. The model then moves back to the left by $16$ tokens and obtains its third segment. The chunking scorer assigns the three segments with the scores $0.24$, $0.87$, and $0.90$, respectively. It suggests that the model considers the third segment as the most informative chunk in answer selection.

\section{Related Work}

There is a growing interest in MRC tasks that require the understanding of both questions and reference documents ~\citep{trischler2017newsqa,rajpurkar2018squad, saeidi2018interpretation,choi2018quac,reddy2018coqa,xu2019review}. Recent studies on pre-trained language models~\citep{radfordimproving,bert2018, DBLP:journals/corr/abs-1907-11769,DBLP:journals/corr/abs-1906-08237} have demonstrated their great success in fine-tuning on MRC tasks. However these pre-trained NLP models (e.g., BERT) only take as input a fixed-length text. Variants of BERT are proposed to process long documents in tasks such as text classification \cite{chalkidis2019neural}.
To deal with lengthy documents in machine reading comprehension tasks, some previous studies skip certain tokens~\citep{yu2017learning,seo2017neural} or select a set of sentences as input based on the given questions~\citep{hewlett2017accurate,min2018efficient,lin2018denoising}. However, they mainly focus on tasks in which most of the answers to given questions are formed by a single informative sentence. These previous approaches are less applicable to deal with those complicated questions that demand cross-sentences reasoning or have much lexical variability from their lengthy documents.

  




\section{Conclusion}
\label{sec:conclusion}
In this paper, we propose a chunking policy network for machine reading comprehension, which enables a model learn to chunk lengthy documents in a more flexible way via reinforcement learning. We also add a recurrent mechanism to allow the information to flow across segments so that the model could have knowledge beyond the current segment when selecting answers. We have performed extensive experiments on three public datasets of machine reading comprehension: CoQA, QuAC, and TriviaQA. Our approach outperforms benchmark models across different datasets.







\bibliography{acl2020.bib}
\bibliographystyle{acl_natbib}

\end{document}